\documentclass{article} 
\usepackage{nips15submit_e,times}
\usepackage{hyperref}
\usepackage{url}

\usepackage{amssymb,amsfonts,amsmath} 
\usepackage{times}
\usepackage{graphicx}
\usepackage{subfigure}
\usepackage{epsfig}
\usepackage{wrapfig}
\usepackage{float}
\usepackage{verbatim}
\usepackage{sidecap}
\usepackage{color}
\usepackage{amsthm}

\def\m#1{\mathbf #1}
\def\bs#1{\boldsymbol{#1}}                  

\def\diag{\textrm{diag}}
\def\diagp#1{\textrm{diag}\big\{#1\big\}}

\def\3D{\mbox{$3$-D}}
\def\2D{\mbox{$2$-D}}
\def\1D{\mbox{$1$-D}}

\newcommand{\ppof}[2]{\dfrac{\partial#1}{\partial#2}}

\newcommand{\?}{\stackrel{?}{=}}

\def\xx{\text{\tiny\raisebox{-0.3em}{\hspace{-0.4em}$xx$}}}
\def\yy{\text{\tiny\raisebox{-0.3em}{\hspace{-0.4em}$yy$}}}
\def\xy{\text{\tiny\raisebox{-0.3em}{\hspace{-0.4em}$xy$}}}
\def\d2{\text{\tiny\raisebox{-0.2em}{\hspace{-0.4em}$\nabla^2$}}}

\title{Uniform Transformation of\\ Non-Separable Probability Distributions}

\author{
Eric Kee\\
\texttt{erickee@gmail.com} \\
}

\nipsfinalcopy 

\begin{document}

\maketitle

\begin{abstract}
A theoretical framework is developed to describe the transformation that distributes probability density functions uniformly over space. In one dimension, the cumulative distribution can be used, but does not generalize to higher dimensions, or non-separable distributions. A potential function is shown to link probability density functions to their transformation, and to generalize the cumulative. A numerical method is developed to compute the potential, and examples are shown in two dimensions. 
\end{abstract}

\section{Introduction}

In a variety of problems, a probability distribution is known, and it is necessary or desirable to compute an explicit mapping between the domain of that distribution, and a secondary domain in which the probability density is distributed uniformly.  A simple example involves drawing samples from a non-analytical, data-specified distribution. With such a mapping, samples can be generated uniformly at random and transformed to produce the data distribution.  In image processing such mappings are used to normalize image brightness (often as a pre-processing stage) to contend with varying illumination conditions~\cite{russ02}. In the field of neural coding, such mappings can be used to compute optimal neural populations for encoding sensory features. 
In the field of optimal coding, such mappings can be used to specify a theoretically optimal companding function. In the context of machine learning, such mappings embody the probabilistic manifold on which the data lie. This work was motivated by manifold learning, and its application to user interface design. An observer may wish to manipulate one or more continuous sliders to navigate a manifold, and such mappings are needed to do so.

In one dimension, the transform has a simple and intuitive solution: the cumulative distribution is computed from the probability density, and used to transform the data. In higher dimensions however, and for non-separable distributions, a solution appears to be missing.  The multidimensional cumulative distribution falls short for non-separable distributions, as each dimension is treated separably~\cite{gersho91}.

The need for a more general understanding of the problem is most clear within the the neural coding literature, which has extensively used the one dimensional cumulative distribution~\cite{ganguli14,ganguli12,wei12,wei12b,ganguli10,mcdonnell08,brunel98,nadal94}.  As a recent example, the cumulative distribution can be used to derive optimal neural populations for one dimensional stimuli~\cite{ganguli14}, thus building a theoretical framework with which to test the efficient coding hypothesis~\cite{barlow61}.  Perceptual stimuli are however multidimensional. Prior work has considered multidimensional stimuli by assuming that distributions are separable, and using the multidimensional cumulative~\cite{nadal94}. Separability is however a strong assumption, and the neural coding literature has noted that a more general solution is needed~\cite{ganguli14,ganguli10}.

In this paper the problem is treated in its general form. We consider how multidimensional, non-separable distributions can be distributed uniformly.  A theoretical framework is developed which shows that a potential function embodies the optimal transformation, and is a generalization of the cumulative.  A numerical method is then developed to compute and visualize the potential and transformation. 

For exposition, Section~\ref{sec:1dtheory} begins with the one dimensional case, and connects the cumulative distribution with a more general theoretical framework that describes the optimal encoding.  The one dimensional theory is then extended to multiple dimensions, and without assuming separability, Section~\ref{sec:multidimensional}.  A potential function is shown to embody the optimal transformation. Section~\ref{sec:manifoldPotential} gives intuition for the shape of the potential, and uses that potential to explain the limitations of the multidimensional cumulative transformation. A numerical method is then developed to compute the potential in two dimensions, Section~\ref{sec:solving2d}.  Lastly, results are presented for a variety of separable and non-separable distributions, Section~\ref{sec:results}, and we conclude in Section~\ref{sec:discussion}.

\section{One Dimensional Distributions}\label{sec:1dtheory}

For exposition, it is useful to begin in one dimension, where a simple solution is known to exist. A function is needed that transforms the domain $x$ of a non-uniform probability density $p(x)$ to a new domain $\hat x$, where its density $u(\hat x)$ is uniform.  Specifically, the transformation must satisfy
\begin{align}\label{eqn:1dstartingpoint_inv}
	\int p(x)dx 
	& ~~=~~ 
	\int u(\hat x)d\hat x~~.
\end{align}
Let $f(x)$ specify how points should be displaced, with $\hat x = x + f(x)$.  Equation~(\ref{eqn:1dstartingpoint_inv}) becomes
\begin{align} \label{eqn:1dchangevariable}
	\int p(x)dx 
	& ~~=~~ 
	\int u(\hat x)\frac{d\hat x}{dx}dx
\\ \label{eqn:integratedconstraint}
	& ~~=~~ 
	\int u(x)(1 + f'(x))dx~~,
\end{align}
where $u(\hat x)$ may be replaced by $u(x)$ to constrain the support of $u$ and $p$ to the same region. Differentiating both sides gives 
\begin{align}\label{eqn:1deqntosolve}
	p/u & ~~=~~
	1 + f'
\end{align}
The derivative of the displacement, $f'$, therefore describes a normalized change in density. The displacement can be recovered by integrating $f' = p/u - 1$, which reveals why the cumulative distribution of $p$ is effective. The displacement $f$ is found by solving a first-order differential equation over normalized probability densities. Computing the cumulative distribution of $p$ does the same, albeit without a normalization to align the support of $p$ and $u$.

\section{Multidimensional Distributions}\label{sec:multidimensional}

In multiple dimensions the cumulative does not generalize because the multidimensional cumulative treats each dimension separably. The theoretical framework described in Section~(\ref{sec:1dtheory}) can however be extended to multiple dimensions without assuming separability. Following the form of Equation~(\ref{eqn:1dstartingpoint_inv}),
\begin{align}
	\int
	p(\m x)d\m x 
	& ~~=~~ 
	\int
	u(\hat{\m x})d\hat{\m x}
\\ \label{eqn:ndformulation}
	& ~~=~~ 
	\int
	u(\m x)| J(\hat{\m x}) |~ d\m x~~,
\end{align}
where $d\m x$ is the volume element, $\hat{\m x} = \m x + \m f(\m x)$ with $\m f$ now specifying a field, and $J$ is the Jacobian. The above follows from the change of variables theorem. The determinant specifies the volume of space after it is warped by the field $\m f$. Taking the derivative gives
\begin{align}\nonumber
	p/u & ~~=~~
	|J(\m x + \m f(\m x))|
\\ \label{eqn:NDjacobianversion}
	& ~~=~~ |\m I + J(\m f) |~~,
\end{align}
which is the multidimensional analog of Equation~(\ref{eqn:1deqntosolve}). This multidimensional expression however leaves the field $\m f$ under-constrained by the scalar change in density, $p/u$.

The field, $\m f$, can be further constrained by noting that it describes only the change in density at each point in space. The change in density under a field is embodied by the field divergence, rather than the curl. Divergence fields are conservative, and can be described as the gradient of a scalar potential function.

Let $g(x)$ be a scalar potential function that embodies the conservative displacement field, $\m f = \nabla g$. Equation~(\ref{eqn:NDjacobianversion}) becomes
\begin{align}\nonumber
	p/u
	& ~~=~~ |\m I + J(\nabla g) |
\\ \label{eqn:NDhessianversion}
	& ~~=~~ |\m I + \m H(g)|~~,
\end{align}
where $\m H$ denotes the Hessian. The transform is therefore $\hat{\m x} = \m x + \nabla g$. This is the multidimensional generalization of the one-dimensional case, Equation~(\ref{eqn:1deqntosolve}). In one dimension, $p / u  = | 1 + g'' | = 1 + f'$.

\section{The Manifold Potential}\label{sec:manifoldPotential}

The potential function $g$, links the density of a distribution $p$ to the displacement field $\m f$ that transforms $p$ to be uniform. Equivalently, $p$ can viewed as a manifold that is non-uniformly distributed through space, and $g$ embodies the points that lie on the manifold. Because that latter geometric interpretation leads to the theoretical framework in Section~(\ref{sec:multidimensional}), we refer to $g$ as the manifold potential of a distribution.

\subsection{Physical Interpretation}

Interpreted physically, the Hessian of the manifold potential describes its local curvature, and it is that curvature which links the probability density $p$ to the displacement field $\m f = \nabla g$, Equation~(\ref{eqn:NDhessianversion}). That local curvature describes how the volume of space is changed by the mapping $\hat{\m x} = \m x + \m f$. Specifically, the Jacobian determinant of the mapping, $|\m I + \m J(\m f)| = |\m I + \m H(g)|$ describes how the curvature of $g$ affects the volume of space.

When the manifold potential has no curvature, the Hessian is the zero matrix, and $p/u = 1$, indicating that $p$ is the uniform distribution. As such, the displacement, $\m f$, should be zero or constant. Because $\m H(g) = \m 0 = J(\m f)$, $\m f$ is zero or constant, as required.

Where the manifold potential has positive curvature, $\m f$ expands $p$ to become less dense. For intuition, consider the case $\m H(g) = \m I$. At such points, $p/u = 2^n$, indicating that $p$ is a factor $2^n$ more dense than the uniform distribution, and must therefore must be rarified. The displacement $\m f$ should expand $p$ to a larger area having twice the size in each dimension, and thus the same density as the uniform distribution. Because the Jacobian determinant of the mapping, $|\m I + \m H(g)| = |2\m I|$, is positive and non-zero, space is expanded as required.

Where the manifold potential has negative curvature, $\m f$ contracts $p$ to become more dense. For intuition, consider the case $\m H(g) = -\alpha \m I$ for $0 < \alpha < 1$. At such points, $p/u = (1-\alpha)^n$, indicating that $p$ is less dense than the uniform distribution, and therefore must be compressed. The displacement $\m f$ must therefore map density from elsewhere within the domain of $p$ to increase the density. Because the Jacobian determinant of the mapping, $|\m I + \m H(g)| = |\m I - \alpha\m I|$, is positive and less than one, space is contracted as required.

Lastly, the manifold potential obeys global physical constraints. The ratio $p/u$, Equation~(\ref{eqn:NDhessianversion}) integrates to $1/u$, the area of the support\footnote{The support of $u$ is defined to encompass $p$, thus $\int p/u = 1/u$ because $\int p = 1$, and $u$ is constant.} of $u$, which constrains the manifold potential. The one dimensional case gives intuition, where the right hand side of Equation~(\ref{eqn:NDhessianversion}) is $|1 + g''|$. We require that $\int (1 + g'') = 1/u$, the area of the support region. Because $\int 1$ is that area, $\int g'' = 0$. The curvature of the one dimensional potential must integrate to zero because it embodies the change in area, and that area must be preserved. That intuition carries into multiple dimensions, where $g''$ is replaced by the Hessian $\m H(g)$, and the change in volume is given by the determinant $|\m I + \m H(g)|$. That change depends upon interactions between the dimensions, and $\m I$ cannot be easily separated as in the one dimensional case. Nonetheless, the curvature of $g$, as embodied by the Hessian, is globally constrained to preserve the volume of the support region.

\subsection{Relationship to the Multidimensional Cumulative}

In a single dimension, the cumulative distribution can be used to transform between a probability distribution $p$ and the uniform distribution $u$. That approach can be generalized into multiple dimensions under the assumption that $p$ is separable~\cite{nadal94}. The multidimensional cumulative distribution provides a transformation. The manifold potential can be used to prove the correctness of that special case, and gives insight into its limitations.

{\em Claim.} For separable distributions which have the form
\begin{align}
	p(\m x) & ~~=~~ \prod_i p_i(x_i)~~,
\end{align}
where $p_i(x_i)$ is a one dimensional distribution over dimension $x_i$ in $\m x$, the probability mass can be transformed to be uniform by
\begin{align}\label{eqn:cumulativeTransform}
	\hat x_i & ~~=~~ a_iP_i(x_i) + b_i~~,
\end{align}
where $P_i$ is the cumulative distribution of $p_i$, and the values $a_i$ and $b_i$ are an unknown affine transform that aligns the support of the uniform distribution to that of $p(\m x)$. 

\begin{proof}
From Section~(\ref{sec:multidimensional}), the transform must satisfy 
\begin{align}
	p/u & ~=~ 	|\m I + \m H(g)|
\\ \label{eqn:transformconstraint}
	\hat{\m x} & ~=~ \m x + \nabla g~~.
\end{align}
Substituting for $p$ gives
\begin{align}\label{eqn:cumulativeEquationToMatch}
	\frac{1}{u}\prod_i p_i(x_i) 
	& ~=~
	|\m I + \m H(g)|~~,
\end{align}
and Equation~(\ref{eqn:transformconstraint}) can be used to compute the elements of $\m I + \m H(g)$:
\begin{align}
	\hat x_i & = x_i + \ppof{g}{x_i} \? a_iP_i(x_i) + b_i
\end{align}
which gives
\begin{alignat}{4}
	\ppof{g}{x_i} & = a_iP_i(x_i) + b_i - x_i~~,
	&\hspace{2em}
	\ppof{^2g}{x_i^2} & = a_ip_i(x_i) - 1~~,
	&\hspace{2em}
	\ppof{^2g}{x_ix_j} & = 0~~.
\end{alignat}
The diagonal elements of $\m I + \m H(g)$, which have the form $\partial^2g/\partial x_i^2$, are therefore $a_ip_i(x_i)$. The off-diagonal elements, which have the form $\partial^2g/\partial x_ix_j$, are zero. The determinant $|\m I + \m H(g)|$ in Equation~(\ref{eqn:cumulativeEquationToMatch}) therefore simplifies,
\begin{align}
	\frac{1}{u}\prod_i p_i(x_i) 
	& ~~=~~
	\prod_i a_ip_i(x_i)~~.
\end{align}
The factors $a_i$ collapse to a single scalar multiplier, which may be defined $\alpha \equiv 1/u$. The transform specified by the multidimensional cumulative distribution is therefore correct for separable distributions up to an affine transform of the domain. 
\end{proof}

The above proof shows that the multidimensional cumulative can be used for separable distributions because the normalized change in density, $p/u$ does not involve interactions between the dimensions, as described by the determinant.  Those interactions are described by the off-diagonal terms in the Hessian of the manifold potential, which the multidimensional cumulative correctly assumes to be zero for separable distributions.

\section{A Numerical Method in Two Dimensions}\label{sec:solving2d}
A method is needed to solve a system of equations having the form $p/u = | \m I + \m H(g)|$. Here the two dimensional case is considered as the simplest, non-trivial instance of the problem. In two dimensions, expanding the determinant gives
\begin{align}\label{eqn:2dcontinuousPDE}
	p/u & ~=~ 
	\ppof{^2g}{x^2} \ppof{^2g}{y^2}
	-\bigg(\ppof{^2g}{xy}\bigg)^{\hspace{-0.2em}2}
	+\ppof{^2g}{x^2} 
	+\ppof{^2g}{y^2}
	+ 1
\end{align}
The potential can be computed over a discrete lattice. Rewriting Equation~(\ref{eqn:2dcontinuousPDE}) in vector form gives
\begin{align}\label{eqn:systemToSolve}
	\bs \rho & ~=~ 
	(\m C_\xx \m g) \circ (\m C_\yy\m g)
	- (\m C_\xy\m g) \circ (\m C_\xy\m g)
	+\m C_\xx\m g
	+\m C_\yy\m g
	~~\equiv~ \m h(\m g)
\end{align}
where $\bs\rho = \m p/\m u - 1$ is the normalized changed in density. The vector $\m g$ comprises samples of $g$ over a \2D lattice. Matrices $\m C$ compute the partial derivatives as discrete convolutions, and $\circ$ denotes the Hadamard product. The function $\m h(\m g)$ is used as shorthand for the above expression.

A gradient based method can be used to solve Equation~(\ref{eqn:systemToSolve}) by minimizing an error function,
\begin{align}\label{eqn:errorfunction}
	E(\m g) & ~=~  \|\m r(\m g)\|^2
\end{align}
where $\m r$ is the residual, $\m r(\m g) = \m h(\m g) - \bs\rho$. The gradient is 
\begin{align}\label{eqn:leastsquaresgradient}
	\nabla E(\m g)
	& ~=~
	2 J[\m h(\m g)]^T
	\m r(\m g)
\end{align}
where $J$ is the Jacobian. By inspection, 
\begin{align}
	J[\m h(\m g)]
	& ~=~
	\diagp{\m C_\xx\m g}\m C_\yy + \diagp{\m C_\yy\m g}\m C_\xx
	-2\diagp{\m C_\xy\m g}\m C_\xy + \m C_\xx + \m C_\yy~~,
\end{align}
where the $\diag\{\}$ operator converts vectors to diagonal matrices. Substituting into Equation~(\ref{eqn:leastsquaresgradient}),
\begin{align}
	\nabla E(\m g)
	& ~=~
	\m C_\yy^T\diagp{\m C_\xx\m g}\m r 
	+ \m C_\xx^T\diagp{\m C_\yy\m g}\m r	
	-2\m C_\xy^T\diagp{\m C_\xy\m g}\m r 
	+ \m C_\xx^T\m r + \m C_\yy^T\m r
\\[0.25em] \label{eqn:errorgradient}
	& ~=~
	\m C_\yy(\m C_\xx\m g \circ \m r)
	+ \m C_\xx(\m C_\yy\m g \circ \m r)		
	-2\m C_\xy(\m C_\xy\m g \circ \m r) 
	+ \m C_\xx \m r + \m C_\yy \m r~~,
\end{align}
where we have used the fact that the convolution matrices are symmetric.  The error gradient can therefore be computed efficiently by separable convolutions.

\subsection{Boundary Constraints}

The potential must obey certain boundary constraints, or probability mass may be mapped outside of its finite domain (mass must be conserved). Specifically, components of $\nabla \m g$ that are orthogonal to the boundary must be zero at the boundary. The potential therefore becomes flat as one approaches the boundary, but may be sloped as one travels along the boundary. That constraint can be enforced implicitly by requiring that the boundary values of $\m g$ be repeated when computing convolutions.

Lastly, a constraint can be added to specify the scalar offset of the potential function. That offset is immaterial, since only the gradient is needed, but can be specified by fixing one boundary value to zero. This detail has not proven necessary in practice, and is omitted for simplicity.

%
\begin{figure}[t]
\begin{center}
   \vspace{-0.5em}
       \subfigure[Ground truth distribution]{
			\includegraphics[width=0.475\linewidth]{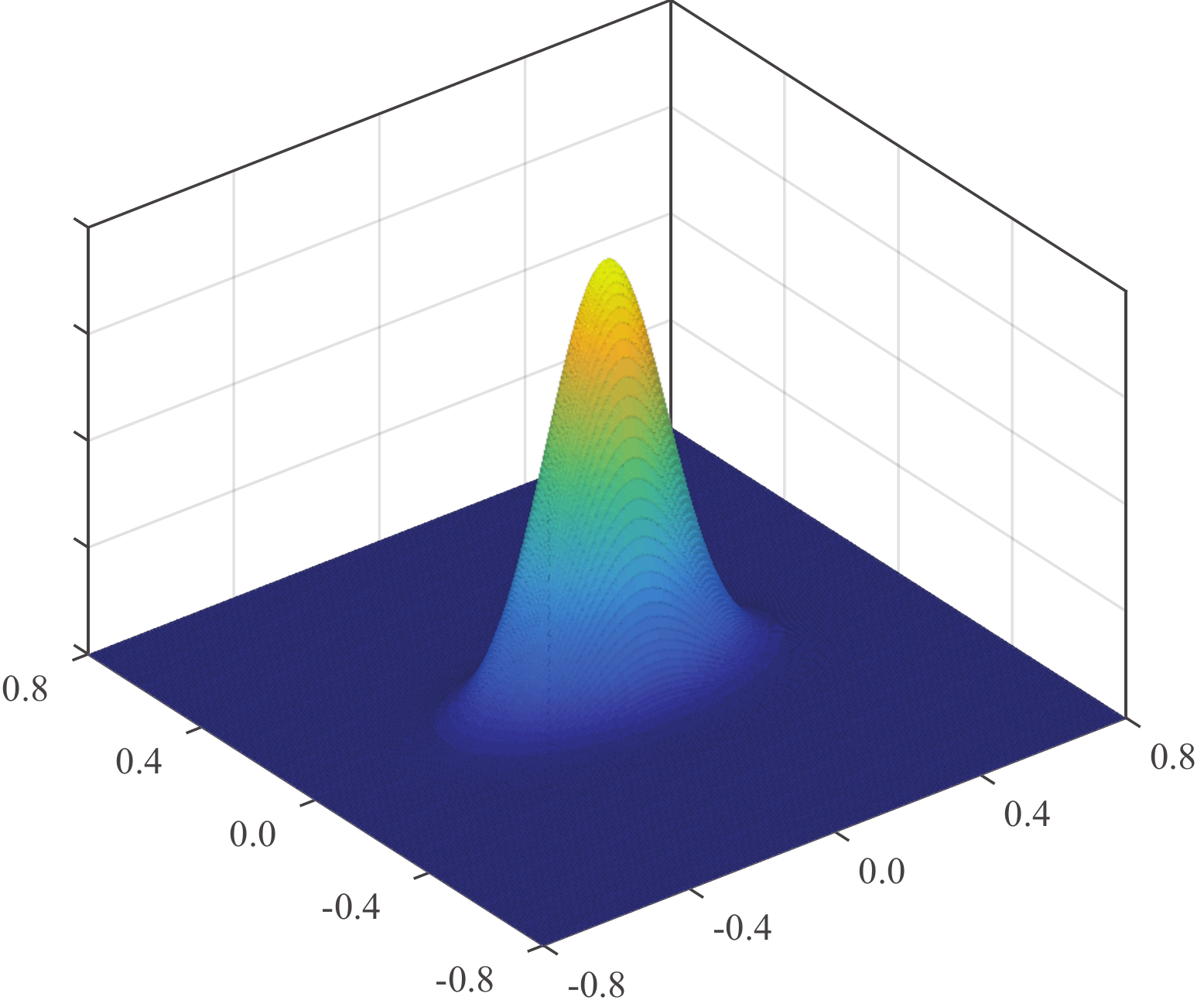}} 
		\hspace{1em}
		\subfigure[Scalar potential $\m g$ and inverse field]{ 
			\includegraphics[width=0.475\linewidth]{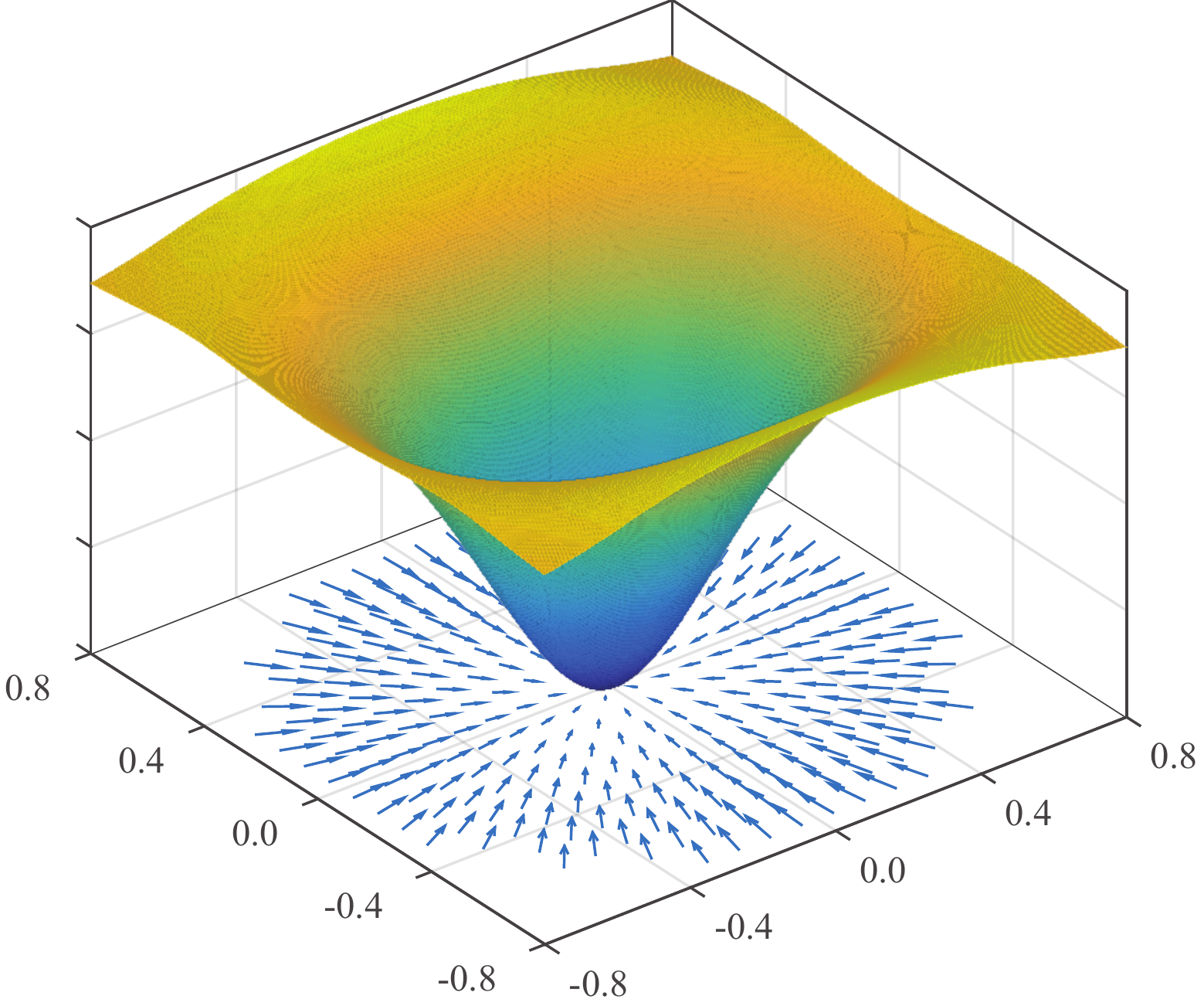}}
\\[-0.5em]
		\hspace{-1.5em}
		\subfigure[Estimated distribution]{
		\includegraphics[width=0.475\linewidth]{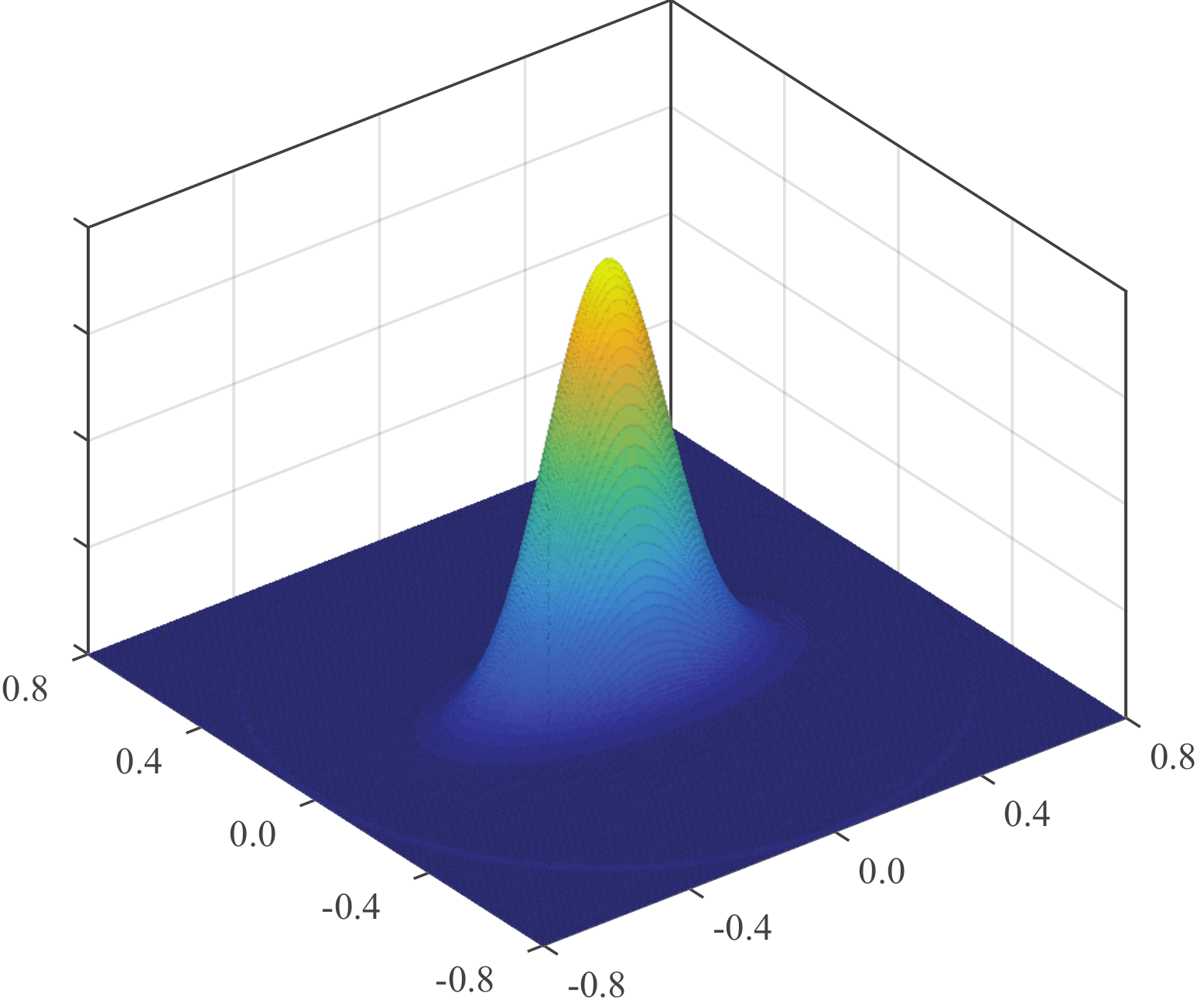}}
		\hspace{2em}
        \subfigure[Contour intervals and inverse field]{
			\includegraphics[width=0.425\linewidth]{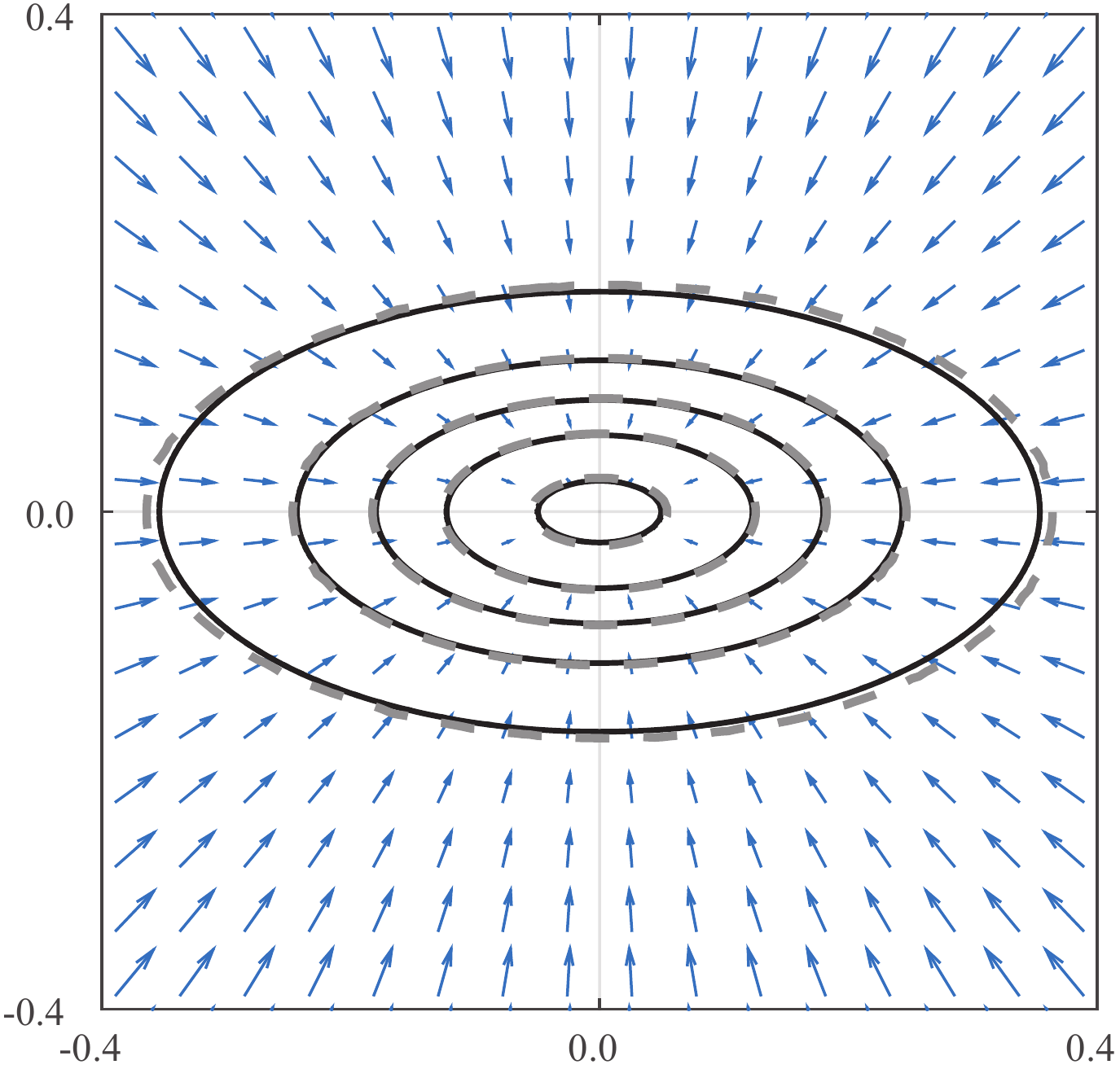}}
    \end{center}
    \vspace{-1.25em}
    \caption{A separable, two dimensional Normal distribution (a); the estimated scalar potential, and its inverse gradient field (b); the estimated distribution (c); and in (d), five contour levels are shown for the ground truth (solid lines) and estimated (dashed) distributions. Vectors are scaled for visualization.}
   \label{fig:gaussian}
   \vspace{-0.5em}
\end{figure}
%

\subsection{Nonlinear Optimization}

Equation~(\ref{eqn:errorfunction}) is solved by a non-linear conjugate gradient method\footnote{Polack-Ribiere method, line search by quadratic and cubic estimation, and Wolfe-Powell stopping criteria.}~\cite{rasmussen10}, with the partials given by Equation~(\ref{eqn:errorgradient}).  Computations are made efficient by separable filters~\cite{farid04}. To ensure convergence, the optimization is wrapped in a multigrid scheme in which a low resolution version of the potential is first estimated. The potential is initialized to zero at the first level.
%
\begin{figure}[t]
\begin{center}
   \vspace{-0.5em}
       \subfigure[Ground truth distribution]{\label{fig:first}
			\includegraphics[width=0.475\linewidth]{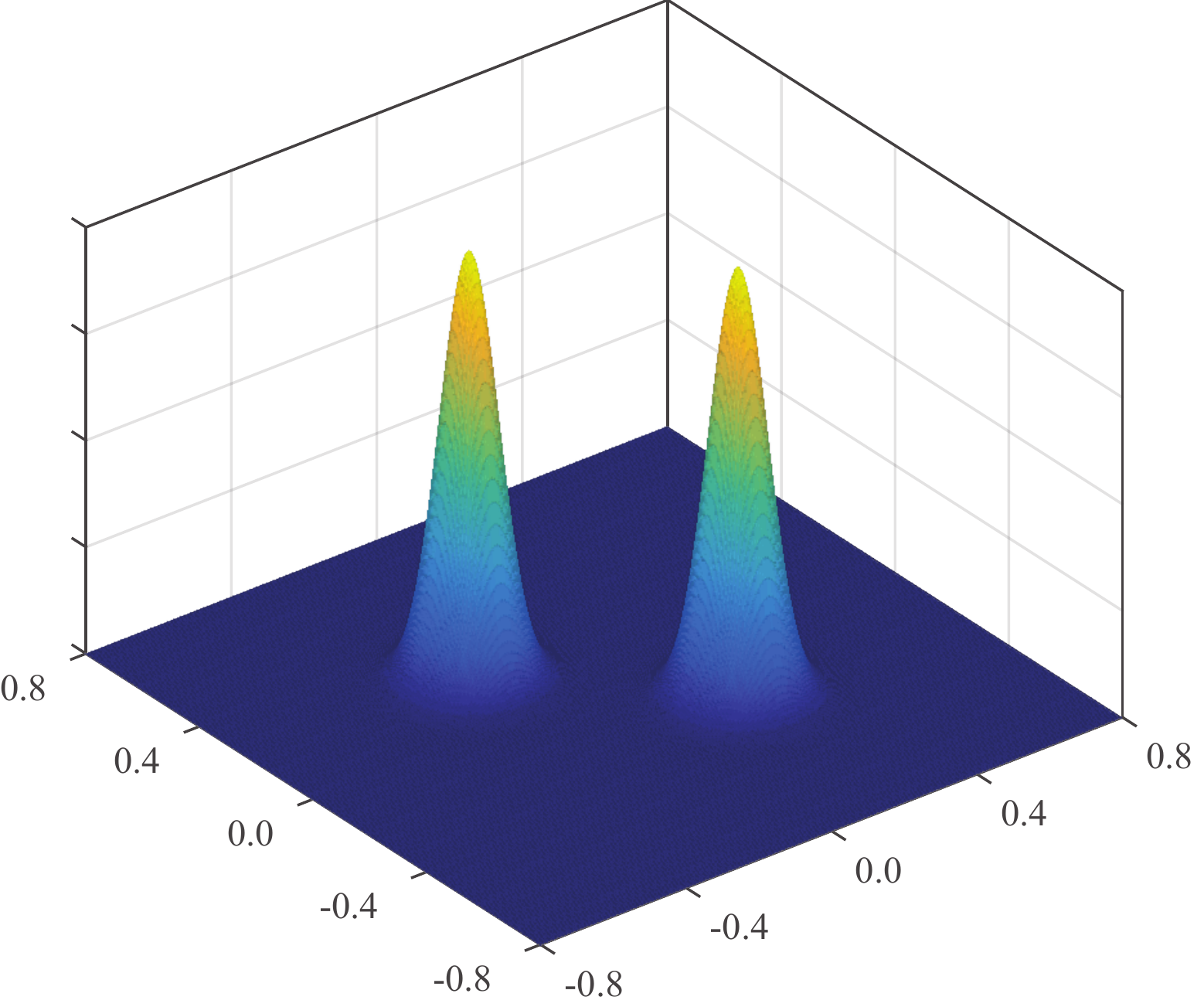}} 
		\hspace{1em}
		\subfigure[Scalar potential $\m g$ and inverse field]{ \label{fig:second}
			\includegraphics[width=0.475\linewidth]{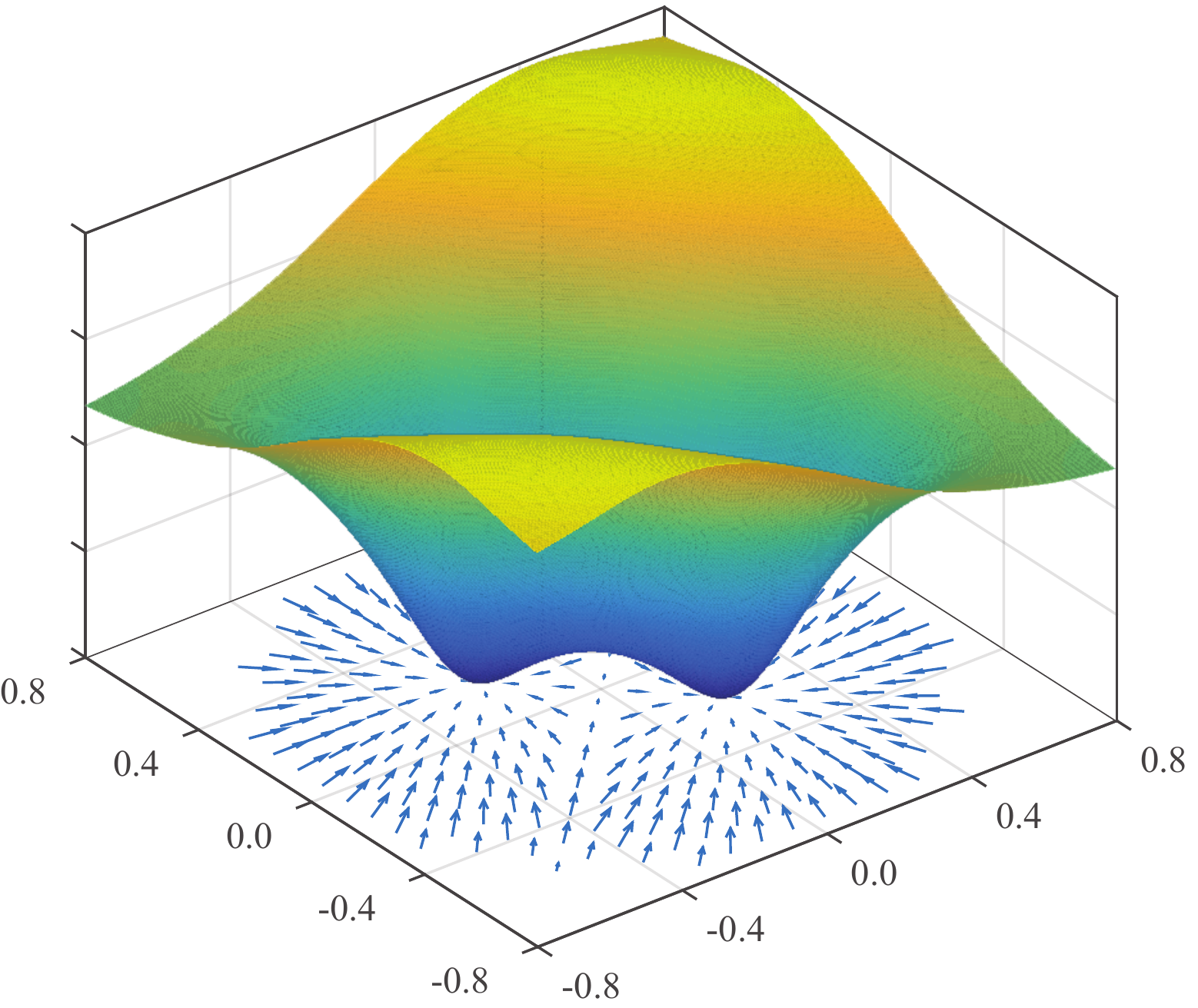}}
\\[-0.5em]
		\hspace{-1.5em}
        \subfigure[Estimated distrubtion]{\label{fig:third}
			\includegraphics[width=0.475\linewidth]{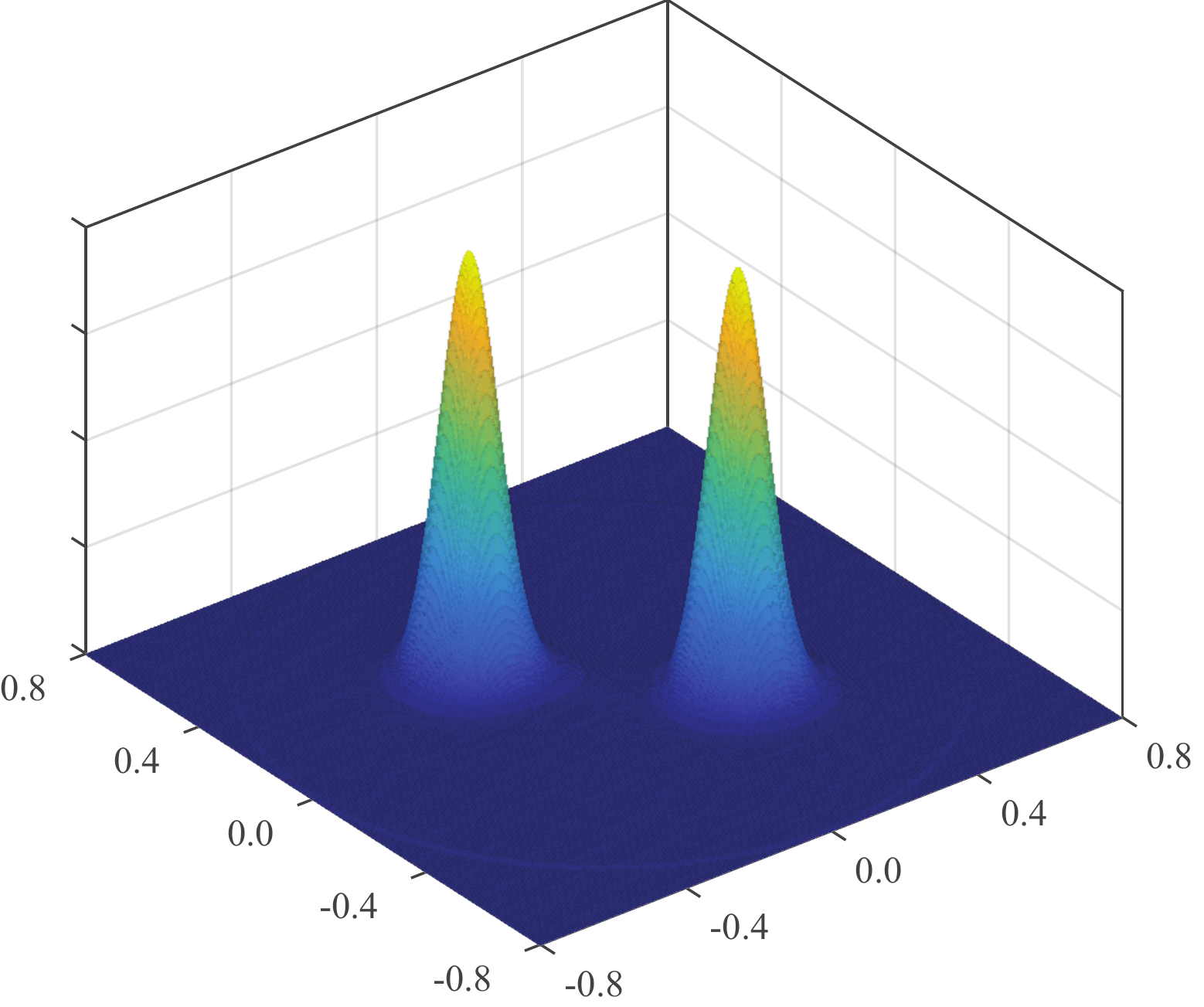}}
		\hspace{2em}
        \subfigure[Contour intervals and inverse field]{\label{fig:fourth}
			\includegraphics[width=0.425\linewidth]{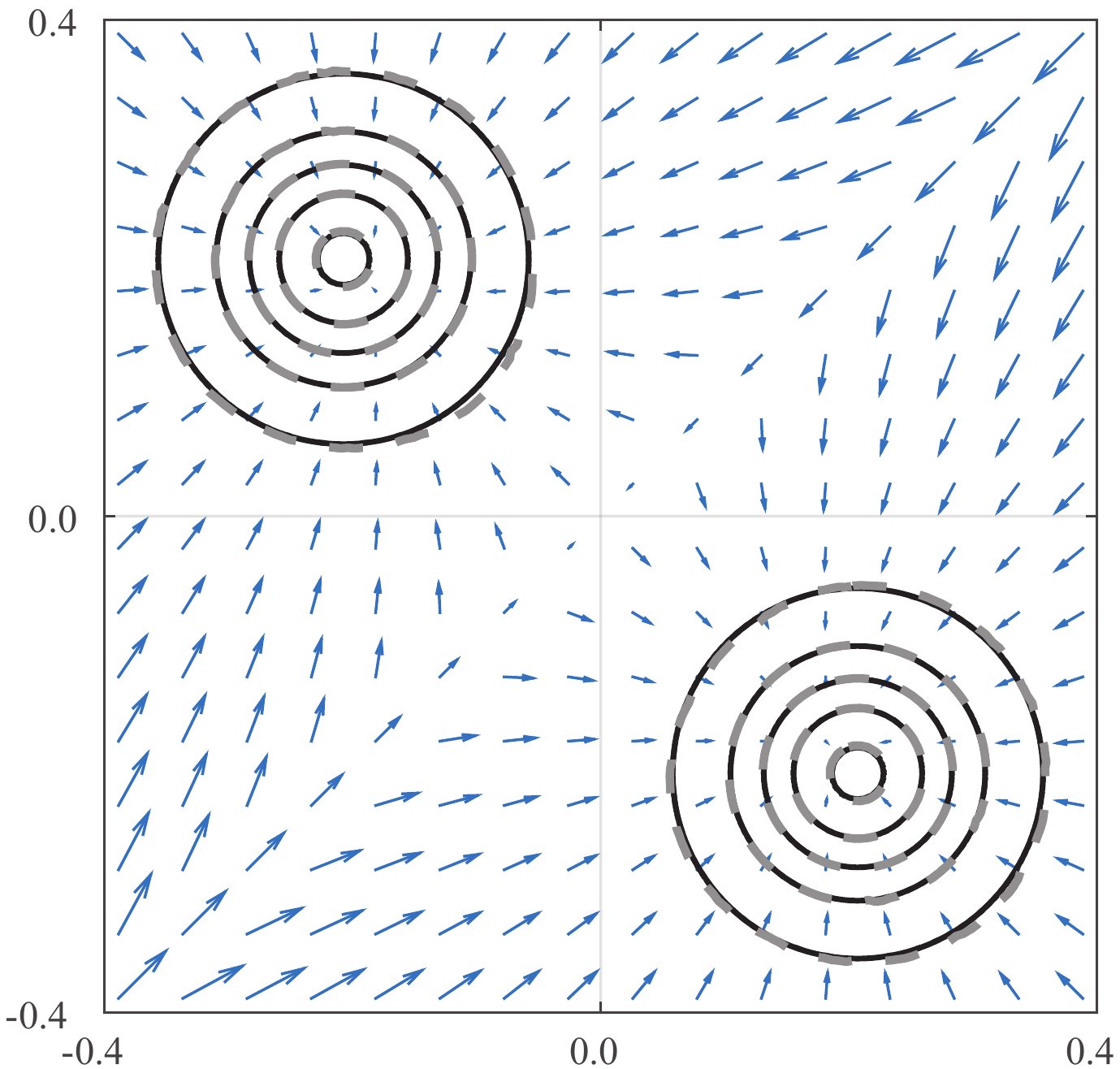}}
    \end{center}
    \vspace{-1.25em}
    \caption{A bimodal Normal distribution (a); the estimated scalar potential, and its inverse gradient field (b); the estimated distribution (c); and in (d), five contour intervals are shown for the ground truth (solid lines) and estimated (dashed) distributions. Vectors are scaled for visualization.}
   \label{fig:bimodal}
   \vspace{-0.75em}
\end{figure}

The above optimization is sufficient when, at the boundary, the probability density is equivalent between the data and uniform distributions ($p = u$ at the boundary). The normalized change in density, $\bs\rho$, is therefore zero on the boundary and beyond. 

For many distributions the density falls to zero at the boundary, while the uniform density does not. The change in density, $\bs \rho$, is therefore nonzero on the boundary (if $p$ falls to zero, $\rho$ falls to $-1$). Outside of the domain, the field is necessarily zero. The potential function is therefore discontinuous at the boundary, and ringing-like artifacts will appear in the numerical solution. The discontinuity in the potential is therefore removed during optimization by windowing $\bs\rho$. Specifically, $\bs\rho$ is padded and circularly windowed to have a sigmoidal transition to zero.  After optimization, the gradient field is cropped accordingly.

\section{Results}\label{sec:results}

To test the numerical method, the multigrid scheme was configured with each level being a factor of two larger than the preceding. The size of the lattice at the smallest level was $44\times 44$ samples, and four levels were used for a maximum resolution of $351\times 351$ samples.  First derivatives were computed using $5$-tap optimal filters~\cite{farid04}, and second derivatives were computed by a second pass\footnote{Two passes afford computation of {\scriptsize $\partial/\partial{xy}$}~\cite{farid04}.}.  The non-linear minimization was configured for $1,000$ line searches at each pyramid level, giving fixed runtimes of $2.6$ minutes\footnote{Implementation is in Matlab running on a modern Macbook Pro.}. The optimization settings were not tuned.
%
\begin{figure}[h]
	\vspace{-0.5em}
	\begin{center}
		\subfigure[Ground truth distribution]{
			\includegraphics[width=0.475\linewidth]{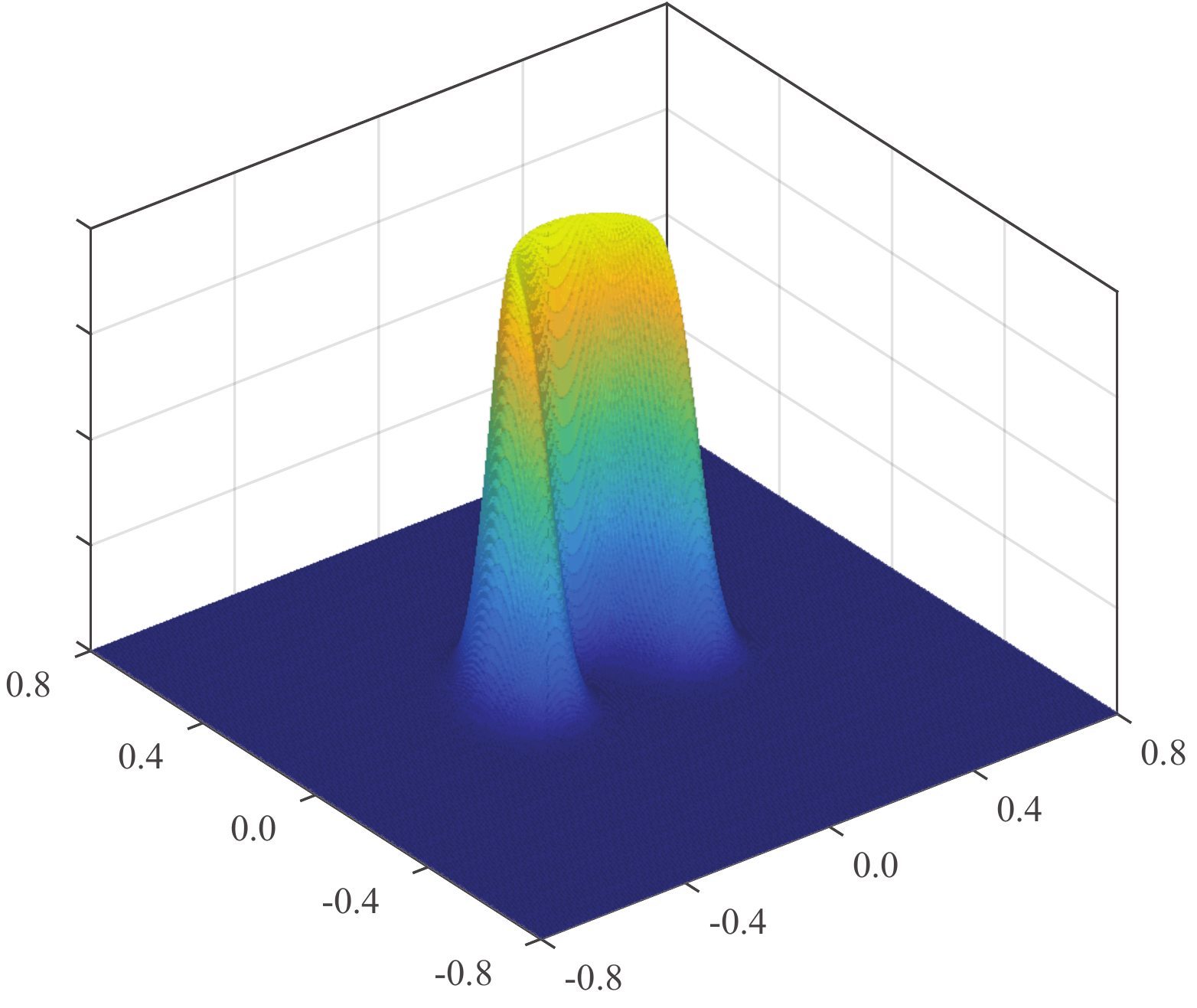}} 
		\hspace{1em}
		\subfigure[Scalar potential $\m g$ and inverse field]{
			\includegraphics[width=0.475\linewidth]{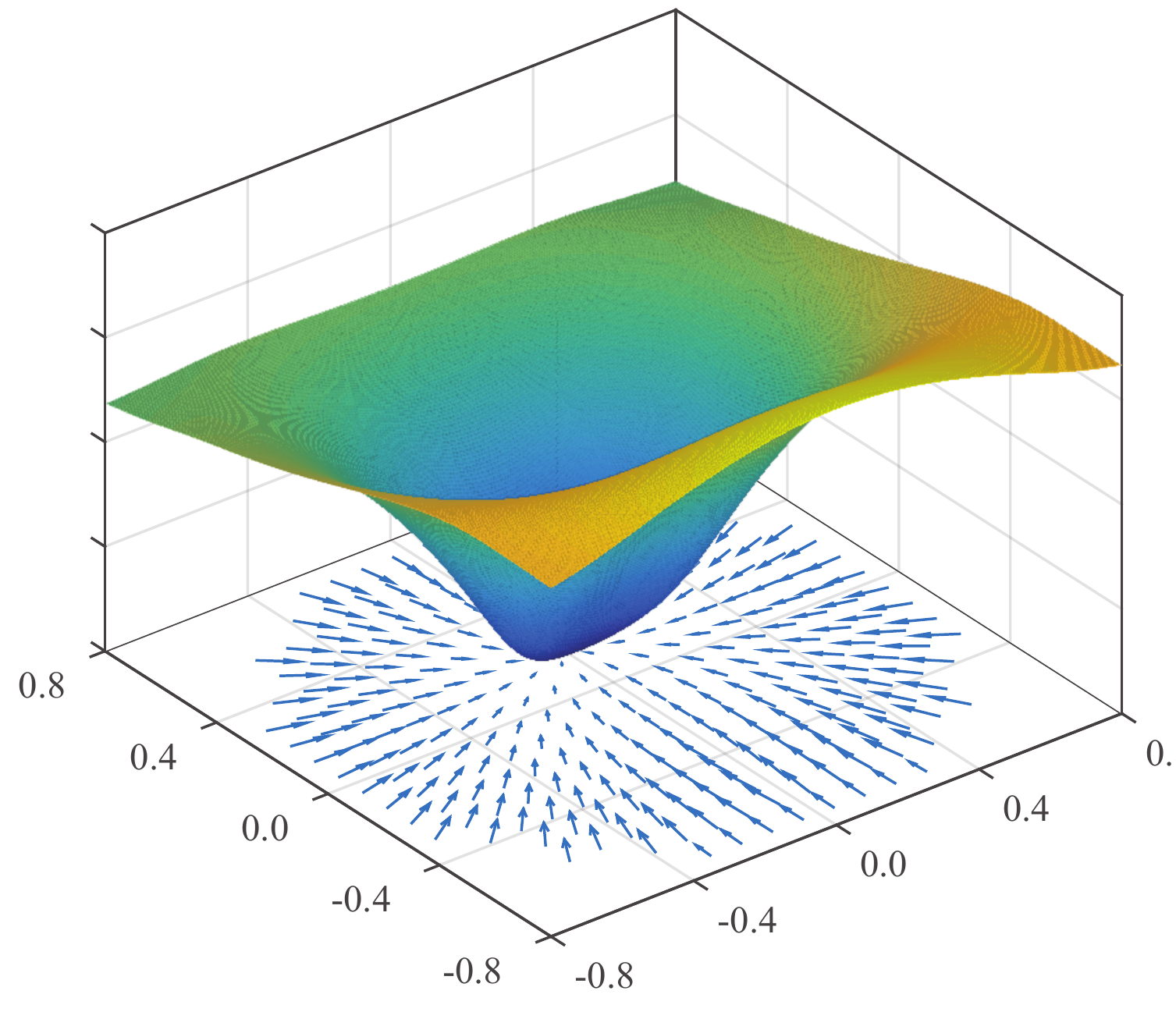}}
		\\[-0.5em]
		\hspace{-1.5em}
		\subfigure[Estimated distribution]{
			\includegraphics[width=0.475\linewidth]{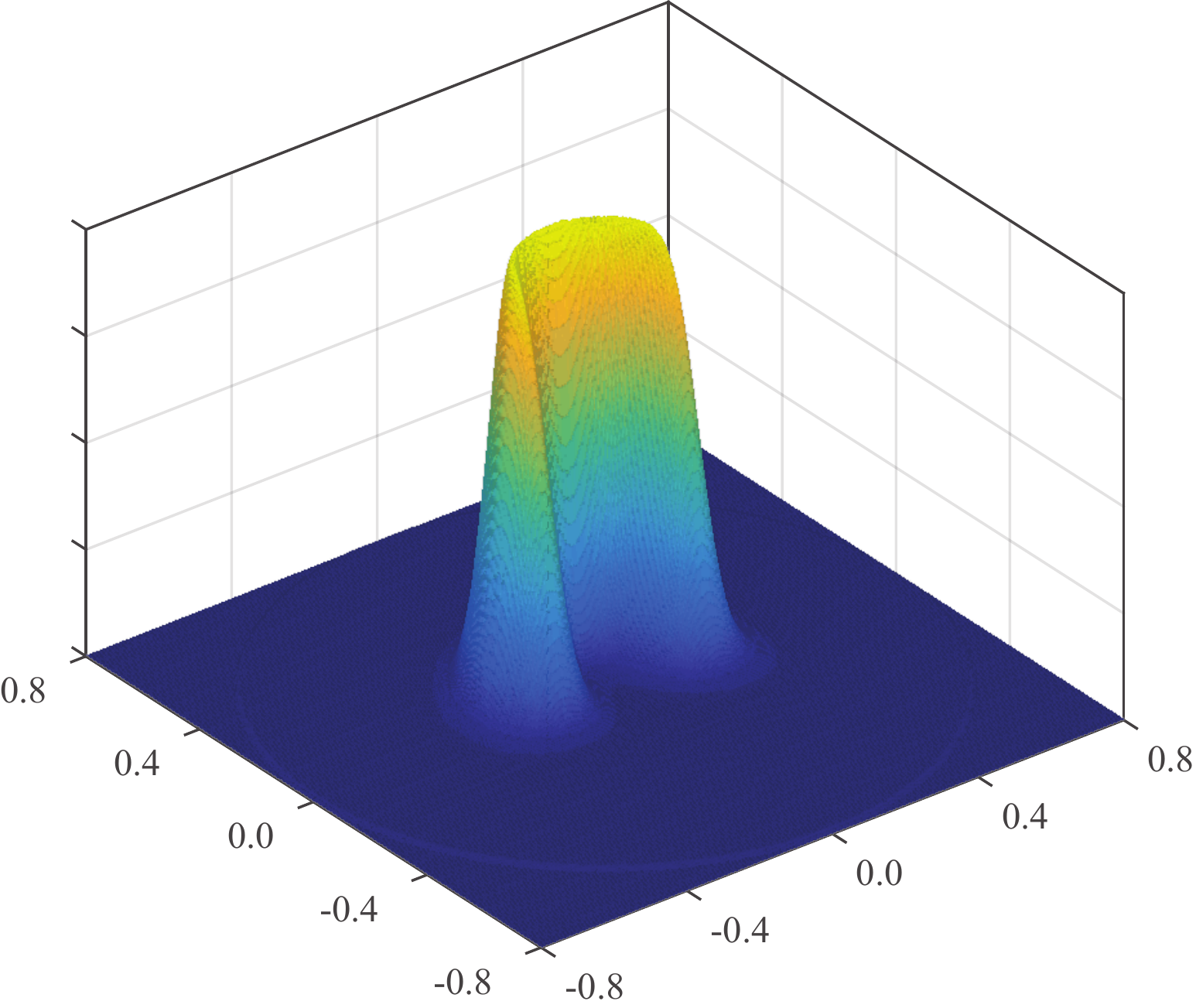}}
		\hspace{2em}
		\subfigure[Contour intervals and inverse field]{
			\includegraphics[width=0.425\linewidth]{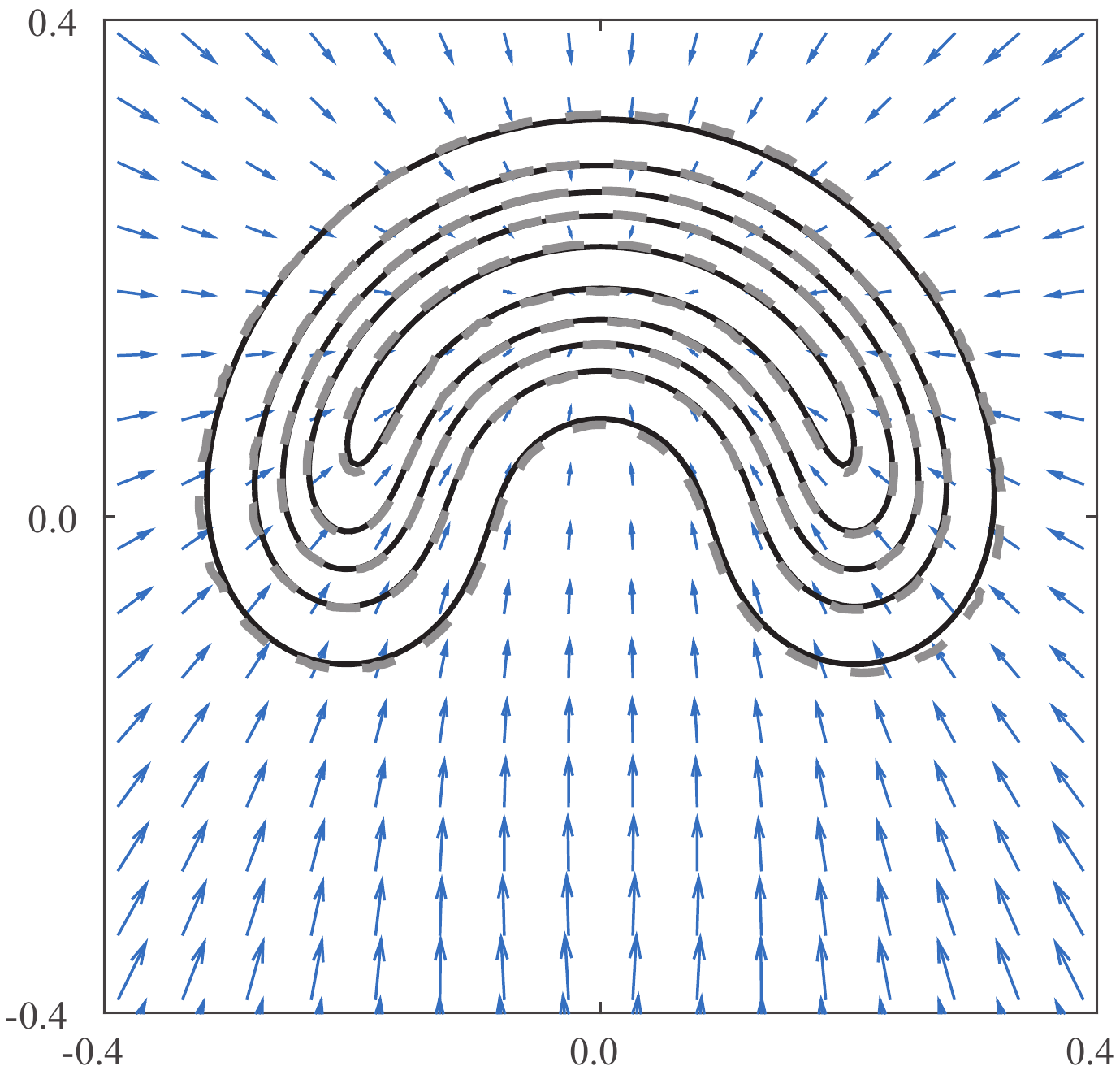}}
	\end{center}
    \vspace{-1.25em}
	\caption{A unimodal distribution with a concavity (a); the estimated scalar potential, and its inverse gradient field (b); the estimated distribution (c); and in (d), five contour levels are shown for the ground truth (solid lines) and estimated (dashed lines) distributions. Vectors are scaled for visualization.
	}
	\label{fig:cshaped}
	\vspace{-0.5em}
\end{figure}

A variety of probability distributions were tested.  To determine how successful was the optimization for each case, the solution field was used to warp a uniform lattice into an estimate of the ground truth probability distribution. Specifically, the inverse of the solution field $\nabla\m g$ was used to warp a uniform lattice. The warped area of each cell was then used to compute the density of the space after warping. Under the correct field, the computed densities should match\footnote{Note that the densities are translated by the field, and must be regridded before comparing to ground-truth.} the ground truth distribution.

Shown in Figure~\ref{fig:gaussian}(a) is a separable, two dimensional Normal distribution.  The potential, $\m g$, and the inverse of its gradient field (b) specify how uniform space is warped to produce the distribution. The field is visualized (blue vectors) below the potential, and within the valid (non-windowed) region of its domain.  Points on a uniform lattice are warped inward toward the center of the distribution.  The magnitude of the field varies\footnote{The asymmetry of the distribution has a subtle effect on the gradient field. The gradient field of a circular Normal distribution forms an ideal radial pattern, whereas an asymmetric Normal distribution produces a field that is (relatively) oblique to the ideal radial orientation.} intuitively according to the shape of the distribution.  Shown in Figure~\ref{fig:gaussian}(c) is the distribution, as reconstructed by warping the uniform lattice.  Contour intervals for both distributions agree closely (d), with solid lines representing the ground truth distribution. The magnitude of the inverse field is scaled for visualization.

The estimated distribution can be quantitatively compared to ground truth by computing the Bhattacharyya coefficient, $\beta = \int\hspace{-0.3em}\sqrt{p(\m x)\hat{p}(\m x)~}d\m x$, which attains a maximum of $1$ when the estimated distribution $\hat{p}$ is identical to ground truth; $0 \le \beta \le 1$.  An intuitive measure of error is therefore the deviation of $\beta$ from $1$. For the separable Normal distribution, Figure~\ref{fig:gaussian}, the error is $0.036$.
%
\begin{figure}[t]
	\vspace{-0.25em}
	\begin{center}
		\subfigure[Ground truth distribution]{\label{fig:first}
			\includegraphics[width=0.475\linewidth]{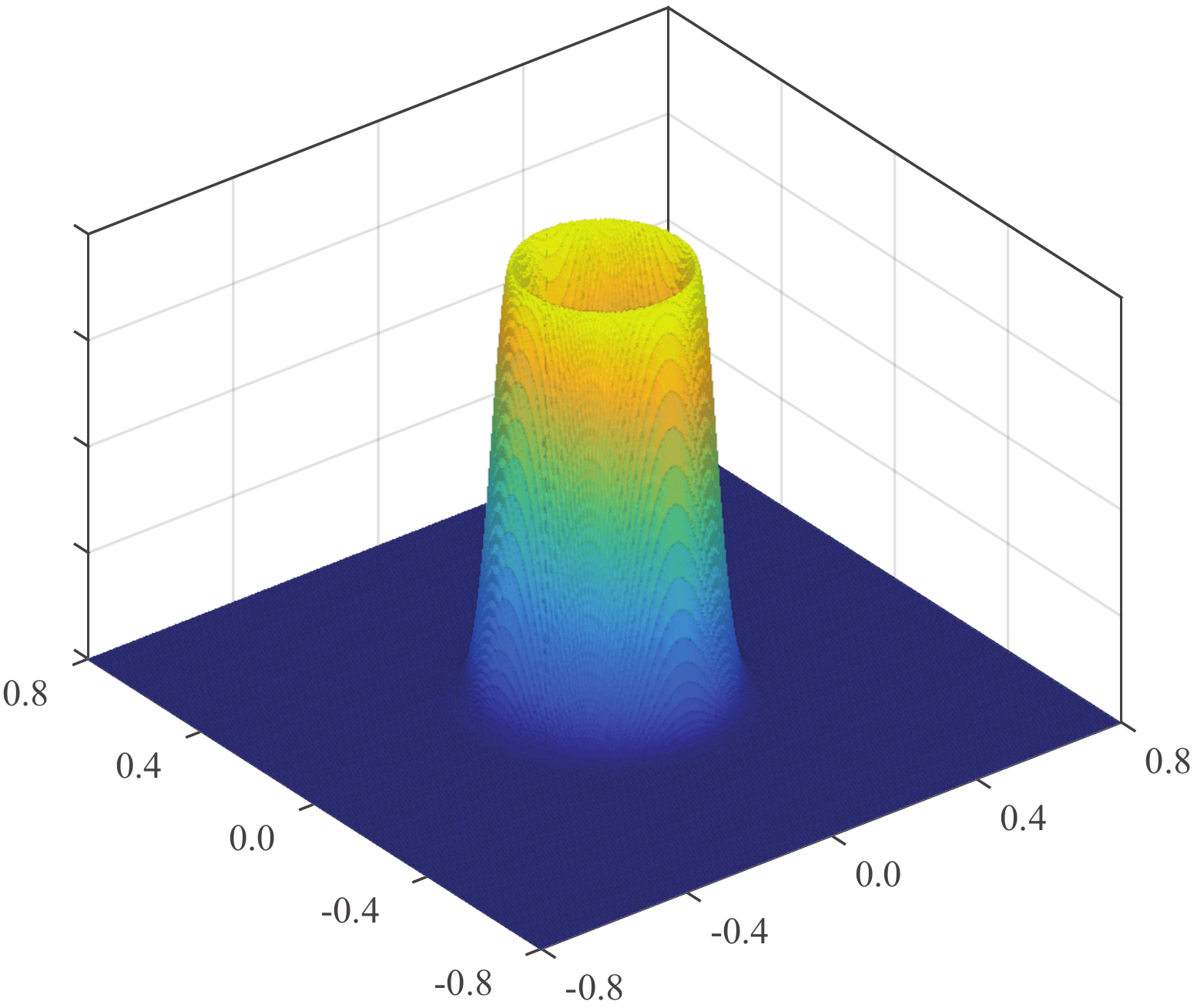}} 
		\hspace{1em}
		\subfigure[Scalar potential $\m g$ and inverse field]{ \label{fig:second}
		\includegraphics[width=0.475\linewidth]{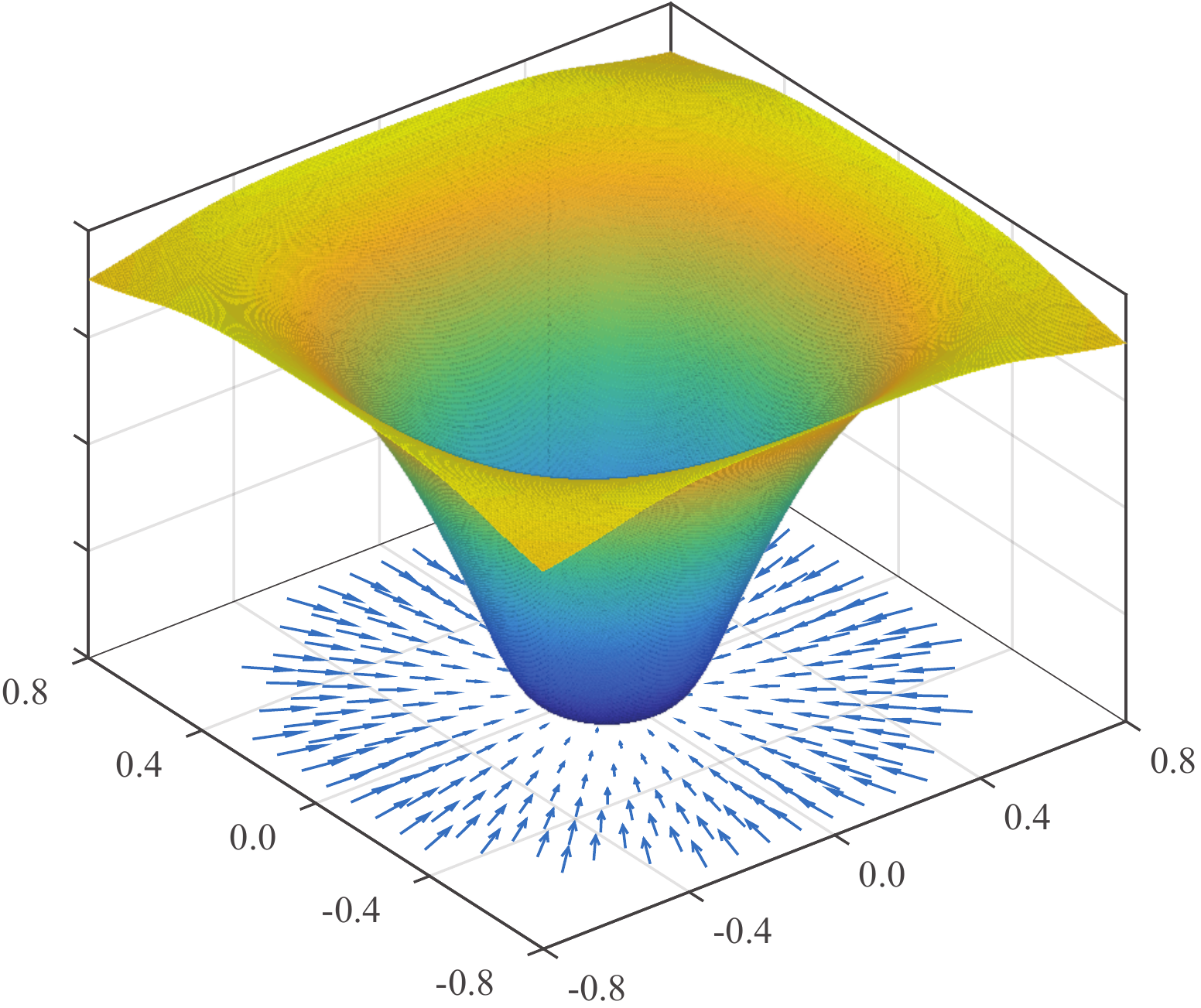}}
		\\[-0.5em]
		\hspace{-1.5em}
		\subfigure[Estimated distribution]{\label{fig:third}
			\includegraphics[width=0.475\linewidth]{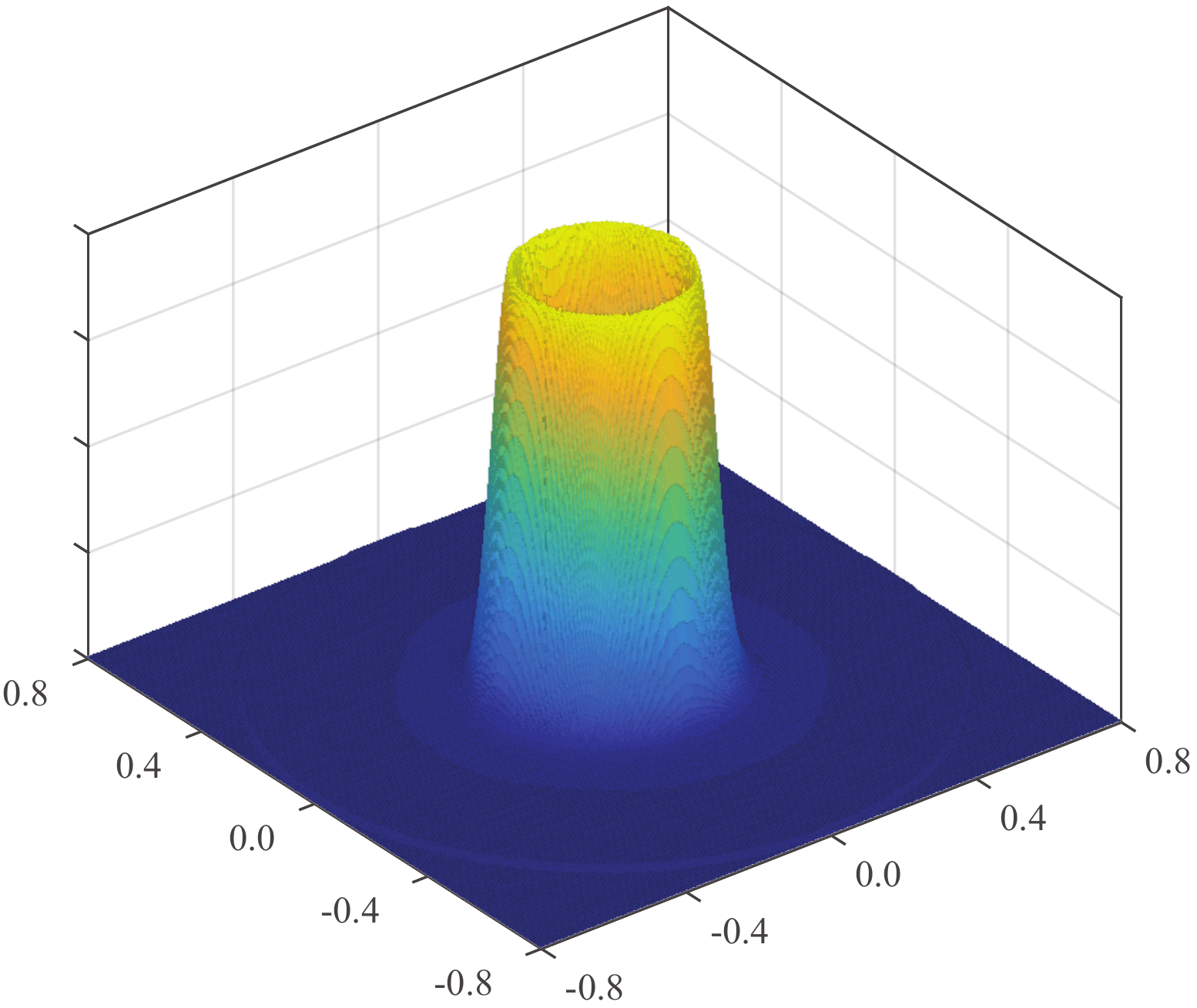}}
		\hspace{2em}
		\subfigure[Contour intervals and inverse field]{\label{fig:fourth}
		\includegraphics[width=0.425\linewidth]{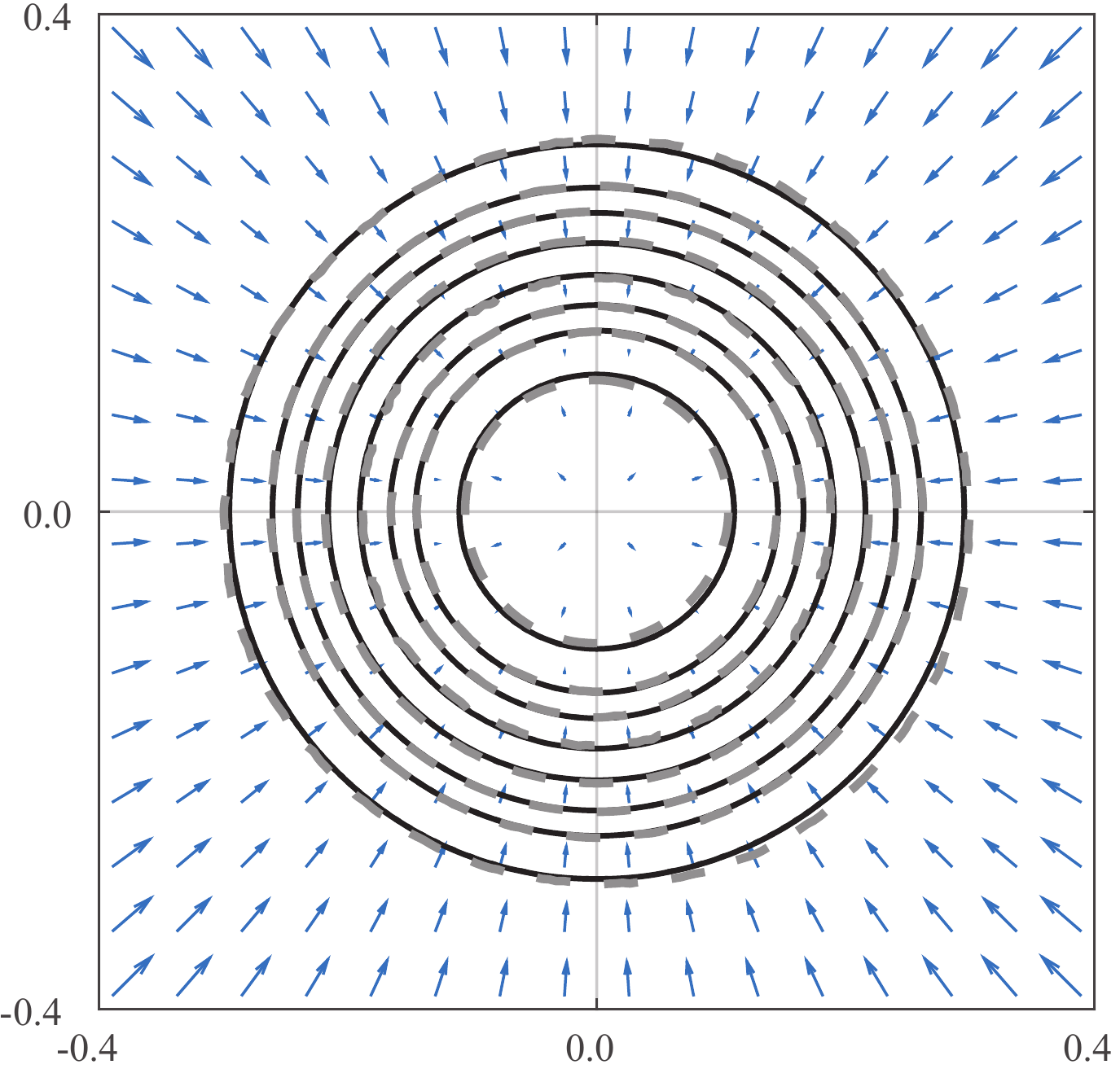}}
	\end{center}
	\vspace{-1.25em}
	\caption{A distribution with a hole (a); the estimated scalar potential, and its inverse gradient field (b); the 	estimated distribution (c); and in (d), four contour levels are shown for the ground truth (solid lines) and estimated (dashed lines) distributions. Vectors are scaled for visualization.
	}
	\label{fig:doughnut}
	\vspace{-0.5em}
\end{figure}

Shown in Figure~\ref{fig:bimodal}(a) is a bimodal Normal distribution. The height of the potential (b) varies around the perimeter, thus affording larger (or smaller) gradient magnitudes where probability mass must be translated further (or closer).  The potential has two distinct minima at the modes of the distribution, and the inverse field (vectors) intuitively warps space toward the modes.  Shown in Figure~\ref{fig:bimodal}(c) is the estimated distribution, and in (c) are the contour intervals. The error is $0.026$.

Shown in Figure~\ref{fig:cshaped}(a) is a unimodal distribution with a concavity. The height of the potential (b) is greater across from the concavity, thus mapping the probability density further.  The inverse gradient field visibly maps probability mass further to produce the concavity.  The estimated distribution (c) and contour intervals (d) align; the error is $0.024$.

Shown in Figure~\ref{fig:doughnut}(a) is a distribution with a hole.  Similar to the Normal distribution, Figure~\ref{fig:gaussian}, the potential is symmetric (b).  The difference can be seen in the slope of the potential, which is steeper near the middle of the domain, and nearly flat at the center.  The steepness results from the more peaked shape of the distribution. The width of the flatter central region corresponds intuitively to the width of the central hole. At the center the potential increases (not visible), producing a field that is oriented outward toward the circular peak.  The estimated distribution (c) and contours intervals (d) match, with $4$ contour levels shown for clarity. The error is $0.023$.

\section{Discussion}\label{sec:discussion}

A theoretical framework was described that defines a mapping between the domain of a uniform distribution, and the domain of non-separable, multidimensional distributions.  The framework was first used to explain why the cumulative distribution uniformly distributes one dimensional distributions, and subsequently extended to an arbitrary number of dimensions. That extension lead to a novel potential function that is associated with probability distributions, and embodies the (probabilistic) manifold. 

The manifold potential describes the relationship between a probability density function, and the field that distributes the probability mass uniformly. The curvature of the potential describes how the density of space should be transformed.  That potential was also shown to explain why the multidimensional cumulative can be used for separable distributions, and why it fails in the non-separable case. The manifold potential therefore generalizes the cumulative by accounting for both separable and non-separable distributions.

Lastly, a numerical method was developed to compute the manifold potential in two dimensions. That gave concrete examples of the manifold potential, and demonstrated that the differential equation is tractable.  The total contribution of this work is therefore a formal and intuitive understanding of the problem, and introduction of a novel concept that is associated with probability distributions.

Minimal assumptions were made about the form of the distributions, requiring primarily that they be discretized for numerical solutions. One limitation is the use of discrete derivatives, which restrict the range of spatial frequencies that can be present in the distribution. Smaller filter widths can partially alleviate this. Nonetheless we anticipate that for many practical cases, distributions will be sufficiently smooth. 

One limitation is the restriction of the numerical method to two dimensions. That restriction was intentional, as it provides results for the smallest non-trivial instance of the problem. The numerical optimization was found to converge reliably. In particular, no significant local minima were observed during the optimizations.  Multiple minima have also not been observed. Numerical methods in higher dimensions are therefore promising. The positive numerical results motivate future work to develop a general numerical method in arbitrary numbers of dimensions. That work is in progress.

To conclude, a broader impact of this work is theoretical. Methods in manifold learning and theories in neuroscience describe encodings for distributions. Those encodings might now be compared to the theoretical optimum that is described by the manifold potential, and novel manifold learning methods might be designed to converge toward the optimum.

\vspace{-0.25em}
{\small \bibliography{arxiv2016}

\begin{thebibliography}{10}

\bibitem{russ02}
Russ, J.~C.,  [{\em Image {P}rocessing
  {H}andbook}{\nolinebreak\hspace{0.1em}]}, CRC Press, Inc., Boca Raton, FL,
  4th~ed. (2002).

\bibitem{gersho91}
Gersho, A. and Gray, R.~M.,  [{\em Vector {Q}uantization and {S}ignal
  {C}ompression}{\nolinebreak\hspace{0.1em}]}, Kluwer Publishers, Norwell, MA
  (1991).

\bibitem{ganguli14}
Ganguli, D. and Simoncelli, E.~P., ``Efficient {S}ensory {E}ncoding and
  {B}ayesian {I}nference with {H}eterogeneous {N}eural {P}opulations,'' {\em
  Neural computation}  (2014).

\bibitem{ganguli12}
Ganguli, D., {\em Efficient {C}oding and {B}ayesian {E}stimation with {N}eural
  {P}opulations}, PhD thesis, New York University (2012).

\bibitem{wei12}
Wei, X.-X. and Stocker, A., ``Efficient {C}oding {P}rovides a {D}irect {L}ink
  {B}etween {P}rior and {L}ikelihood in {P}erceptual {B}ayesian {I}nference,''
  in [{\em Advances in Neural Information Processing Systems
  25}{\nolinebreak\hspace{0.1em}]},  (2012).

\bibitem{wei12b}
Wei, X.-X. and Stocker, A., ``Bayesian {I}nference with {E}fficient {N}eural
  {P}opulation {C}odes,'' in [{\em Artificial Neural Networks and Machine
  Learning}{\nolinebreak\hspace{0.1em}]},  {\em Lecture Notes in Computer
  Science} {\bf 7552}, Springer Berlin Heidelberg (2012).

\bibitem{ganguli10}
Ganguli, D. and Simoncelli, E.~P., ``Implicit {E}ncoding of {P}rior
  {P}robabilities in {O}ptimal {N}eural {P}opulations,'' in [{\em Advances in
  Neural Information Processing Systems 23}{\nolinebreak\hspace{0.1em}]},
  (2010).

\bibitem{mcdonnell08}
McDonnell, M.~D. and Stocks, N.~G., ``Maximally {I}nformative {S}timuli and
  {T}uning {C}urves for {S}igmoidal {R}ate-{C}oding {N}eurons and
  {P}opulations,'' {\em Physical Review Letters}~{\bf 101} (2008).

\bibitem{brunel98}
Brunel, N. and Nadal, J.-P., ``Mutual {I}nformation, {F}isher {I}nformation,
  and {P}opulation {C}oding,'' {\em Neural Computation}~{\bf 10}(7) (1998).

\bibitem{nadal94}
Nadal, J.-P. and Parga, N., ``Nonlinear {N}eurons in the {L}ow-{N}oise {L}imit:
  {A} {F}actorial {C}ode {M}aximizes {I}nformation {T}ransfer,'' {\em Network:
  Computation in Neural Systems}~{\bf 5}(4) (1994).

\bibitem{barlow61}
Barlow, H.~B.,  [{\em Possible {P}rinciples {U}nderlying the {T}ransformation
  of {S}ensory {M}essages}{\nolinebreak\hspace{0.1em}]}, MIT Press, Cambridge,
  MA (1961).

\bibitem{rasmussen10}
Rasmussen, C.~E. and Nickisch, H., ``Gaussian {P}rocesses for {M}achine
  {L}earning ({GPML}) {T}oolbox,'' {\em Journal of Machine Learning
  Research}~{\bf 11} (2010).

\bibitem{farid04}
Farid, H. and Simoncelli, E., ``Differentiation of {D}iscrete
  {M}ultidimensional {S}ignals,'' {\em IEEE Transactions on Image
  Processing}~{\bf 13}(4) (2004).

\end{thebibliography}
\bibliographystyle{spiebib}}

\end{document}